\documentclass[conference]{IEEEtran}
\IEEEoverridecommandlockouts
\usepackage{epsfig}
\usepackage{times}
\usepackage{float}
\usepackage{afterpage}
\usepackage{amsmath}
\usepackage{amssymb}
\usepackage{amstext}
\usepackage{bm}
\usepackage{latexsym}
\usepackage{color}
\usepackage{array,algpseudocode}
\usepackage{makecell}
\usepackage{diagbox}
\usepackage{amsthm}
\usepackage{amssymb} 
\usepackage{graphicx}
\usepackage{caption}
\usepackage{pstricks}
\usepackage{booktabs}
\usepackage{enumerate}
\usepackage[ruled]{algorithm}
\usepackage{cite}
\usepackage{subcaption}
\usepackage{flushend}
\usepackage[left = 0.72in, right=0.72in, top=0.75in, bottom=1.05in]{geometry}
\usepackage{bbm}

\usepackage[normalem]{ulem}
\usepackage{url}

\newcommand{\comment}[1]{}

\usepackage{mathbbol}

\def\mindex#1{\index{#1}}

\def\sq{\hbox{\rlap{$\sqcap$}$\sqcup$}}
\def\qed{\ifmmode\sq\else{\unskip\nobreak\hfil
\penalty50\hskip1em\null\nobreak\hfil\sq
\parfillskip=0pt\finalhyphendemerits=0\endgraf}\fi\medskip}

\long\def\defbox#1{\framebox[.9\hsize][c]{\parbox{.85\hsize}{%
\parindent=0pt
\baselineskip=12pt plus .1pt      
\parskip=6pt plus 1.5pt minus 1pt 
 #1}}}

\long\def\beginbox#1\endbox{\subsection*{}%
\hbox{\hspace{.05\hsize}\defbox{\medskip#1\bigskip}}%
\subsection*{}}

\def\endbox{}

\newsavebox{\junk}
\savebox{\junk}[1.6mm]{\hbox{$|\!|\!|$}}

\def\bfmath#1{{\mathchoice{\mbox{\boldmath$#1$}}%
{\mbox{\boldmath$#1$}}%
{\mbox{\boldmath$\scriptstyle#1$}}%
{\mbox{\boldmath$\scriptscriptstyle#1$}}}}

\def\bfmY{\bfmath{Y}}

\def\bfmhhaY{\bfmath{\hhaY}} 
\def\bfmhhaY{\hbox to 0pt{$\widehat{\bfmY}$\hss}\widehat{\phantom{\raise 1.25pt\hbox{$\bfmY$}}}}

\def\til={{\widetilde =}}

 \def\FRAC#1#2#3{\genfrac{}{}{}{#1}{#2}{#3}}

\def\ddtp{{\mathchoice{\FRAC{1}{d^{\hbox to 2pt{\rm\tiny +\hss}}}{dt}}%
{\FRAC{1}{d^{\hbox to 2pt{\rm\tiny +\hss}}}{dt}}%
{\FRAC{3}{d^{\hbox to 2pt{\rm\tiny +\hss}}}{dt}}%
{\FRAC{3}{d^{\hbox to 2pt{\rm\tiny +\hss}}}{dt}}}}

\def\average#1,#2,{{1\over #2} \sum_{#1}^{#2}}

\def\eye(#1){{\bf(#1)}\quad}

\def\eq#1/{(\ref{e:#1})}

\newcommand{\beqn}[1]{\notes{#1}%
\begin{eqnarray} \elabel{#1}}

\newcommand{\eeqn}{\end{eqnarray} }

\newcommand{\beq}[1]{\notes{#1}%
\begin{equation}\elabel{#1}}

\newcommand{\eeq}{\end{equation}}

\def\bdes{\begin{description}}
\def\edes{\end{description}}

\newcounter{rmnum}

\newcounter{anum}

{\end{list}}

\def\ass(#1:#2){(#1\ref{#1:#2})}

\def\ritem#1{
\item[{\sf \ass(\current_model:#1)}]
}

\newenvironment{recall-ass}[1]{%
\begin{description}
\def\current_model{#1}}{
\end{description}
}

\long\def\comment#1{}

\newfont{\bb}{msbm10 scaled 1100}

\let\svthefootnote\thefootnote
\newcommand\blankfootnote[1]{%
  \let\thefootnote\relax\footnotetext{#1}%
  \let\thefootnote\svthefootnote%
}

\allowdisplaybreaks

\begin{document}
\title{Echo State Network (ESN) for Signal Recovery in RF-Impaired IBFD MIMO Systems
\thanks{This work was supported in part by the Department of Transportation (DOT) Tier-1 University Transportation Center for Advancing Cybersecurity Research and Education (CYBER-CARE), and in part by the Embry-Riddle Aeronautical University Faculty Innovation Research in Science and Technology (FIRST) program. }
}
\author{

\IEEEauthorblockN{
Conrad Prisby\IEEEauthorrefmark{1},
Siyao Li\IEEEauthorrefmark{1},
Chengtao Xu\IEEEauthorrefmark{2},
Thomas Yang\IEEEauthorrefmark{1}
}

\IEEEauthorblockA{
\IEEEauthorrefmark{1}
Department of Electrical Engineering and Computer Science,\\
Embry-Riddle Aeronautical University, Daytona Beach, FL, USA
}

\IEEEauthorblockA{
\IEEEauthorrefmark{2}
The Johns Hopkins University Applied Physics Laboratory,
Laurel, MD, USA
}

\IEEEauthorblockA{
E-mails:
\{prisbyc@my.erau.edu,
lis14@erau.edu,
chengtao.xu@jhuapl.edu,
yang482@erau.edu\}
}

}

\maketitle

\begin{abstract}

In-band full-duplex (IBFD) multiple-input multiple-output (MIMO) systems enable simultaneous transmission and reception on the same frequency band, improving spectral efficiency for next-generation wireless networks. However, IBFD-MIMO systems are susceptible to self-interference (SI), which may overpower signals of interest (SOI). In this scenario, blind source separation (BSS) algorithms can be adopted to remove SI and perform joint sensing and communication (JSAC), but BSS algorithms mostly assume an idealized linear and quasi-stationary signal model, which does not hold under realistic radio frequency (RF) impairments, such as I/Q imbalance, carrier frequency offset (CFO), phase noise, and power amplifier nonlinearity. This paper proposes a two-stage echo state network (ESN)-based scheme that is superior to BSS under these realistic conditions. A frozen ESN is trained offline to characterize the static SI path, while an adaptive ESN, updated online via recursive least squares, tracks the time-varying SOI path using sparse pilot symbols. We evaluate the proposed scheme's SOI recovery performance and acquisition speed with different block sizes, comparing it against other recurrent neural networks (RNN), such as long short-term memory (LSTM) and gated recurrent unit (GRU). Simulation results show that the proposed approach outperforms BSS, LSTM, and GRU in both efficiency and SOI recovery, demonstrating the viability of ESNs for real-time, nonlinear self-interference cancellation in realistic IBFD MIMO systems.

\begin{IEEEkeywords}
Echo state networks, MIMO, self-interference cancellation, RF impairments, recursive least squares.
\end{IEEEkeywords}

\end{abstract}

\section{Introduction}
\label{sec:intro}

The growth in mobile data traffic and the expansion into millimeter-wave (mmWave) frequencies have increased the demand for higher spectral efficiency schemes in next-generation wireless networks \cite{heath2016overview}, where the overhead of acquiring and tracking directional channel state becomes a fundamental limit on achievable rate~\cite{Li2022,Li2024}.
To address the spectrum scarcity issue, in-band full-duplex (IBFD) multiple-input multiple-output (MIMO) systems have gained popularity, since they allow transmission and reception to occur simultaneously on the same frequency band \cite{kolodziej2021band, alves2020full}. However, IBFD systems suffer from self-interference (SI), where a node’s own transmission leaks into its receive chain and can overpower the signal of interest (SOI).

The SI challenge can be addressed by applying blind source separation (BSS) algorithms, which are capable of jointly recovering the SOI and estimating the reflected SI channel. 
Our previous work used fast independent component analysis (Fast-ICA) algorithm for BSS in IBFD MIMO systems to perform joint sensing and communication (JSAC), leveraging the knowledge of the SI signal for self-interference cancellation (SIC)\cite{Li2025MILCOM, Li2025Electronics}.  
Those studies exposed a block-size trade-off under time-varying channels, with SOI recovery degrading for large blocks. 
This trade-off mirrors the one arising in channels with block memory and feedback, where the optimal split between channel-state acquisition and data transmission has been characterized information-theoretically for both binary~\cite{Li2023} and Gaussian~\cite{Li2024} ``beam-pointing'' models.
However, these results assumed linear mixing and real-valued symbols without transmitter or receiver RF impairments. 
In-phase/quadrature (I/Q) imbalance introduces conjugate coupling, oscillator offsets rotate the constellation, and power-amplifier
(PA) nonlinearity distorts symbol amplitudes and phases~\cite{Wang_CrystalOscillators}. PA memory and multipath further introduce temporal dependence \cite{Anttila2013Cancellation, Ding2004Memory}.
These effects motivate receiver models beyond the
instantaneous linear mixing assumed by Fast-ICA.

An echo state network (ESN) is a well-suited adaptive approach because it is a type of recurrent neural network (RNN) that uses  fixed input and recurrent weights with a trained readout \cite{9966815}. 
Reservoir computing (RC) is significantly more computationally efficient than conventional RNNs like long short-term memory (LSTM) and gated recurrent unit (GRU) \cite{LSTM, GRU}. 
Mosleh et al. \cite{8169663} studied RC-based MIMO-OFDM detection under nonlinear distortion. Subsequent work by Li et al.~\cite{li2023real} developed online multi-mode RC detection with alternating RLS updates, decision feedback, and nonlinear compensation.
These methods establish online reservoir detection and RLS adaptation as building blocks for the receiver considered here.
For SI suppression, Liu et al.~\cite{10556632} developed RC-based digital SIC and compared cancellation performance and complexity against polynomial and neural-network cancellers. Memory-polynomial cancellers can model nonlinear distortion with memory, but require the nonlinear order and memory depth to be selected in advance, with the number of model terms increasing with both parameters. In contrast, an ESN produces nonlinear features through its activation function and captures temporal dependencies through its recurrent state, while requiring only the output layer to be trained. This structure supports efficient offline calibration or online adaptation through linear methods such as ridge regression and RLS.

\begin{figure}[!ht]
    \centerline{ \includegraphics[width=1.00\linewidth]{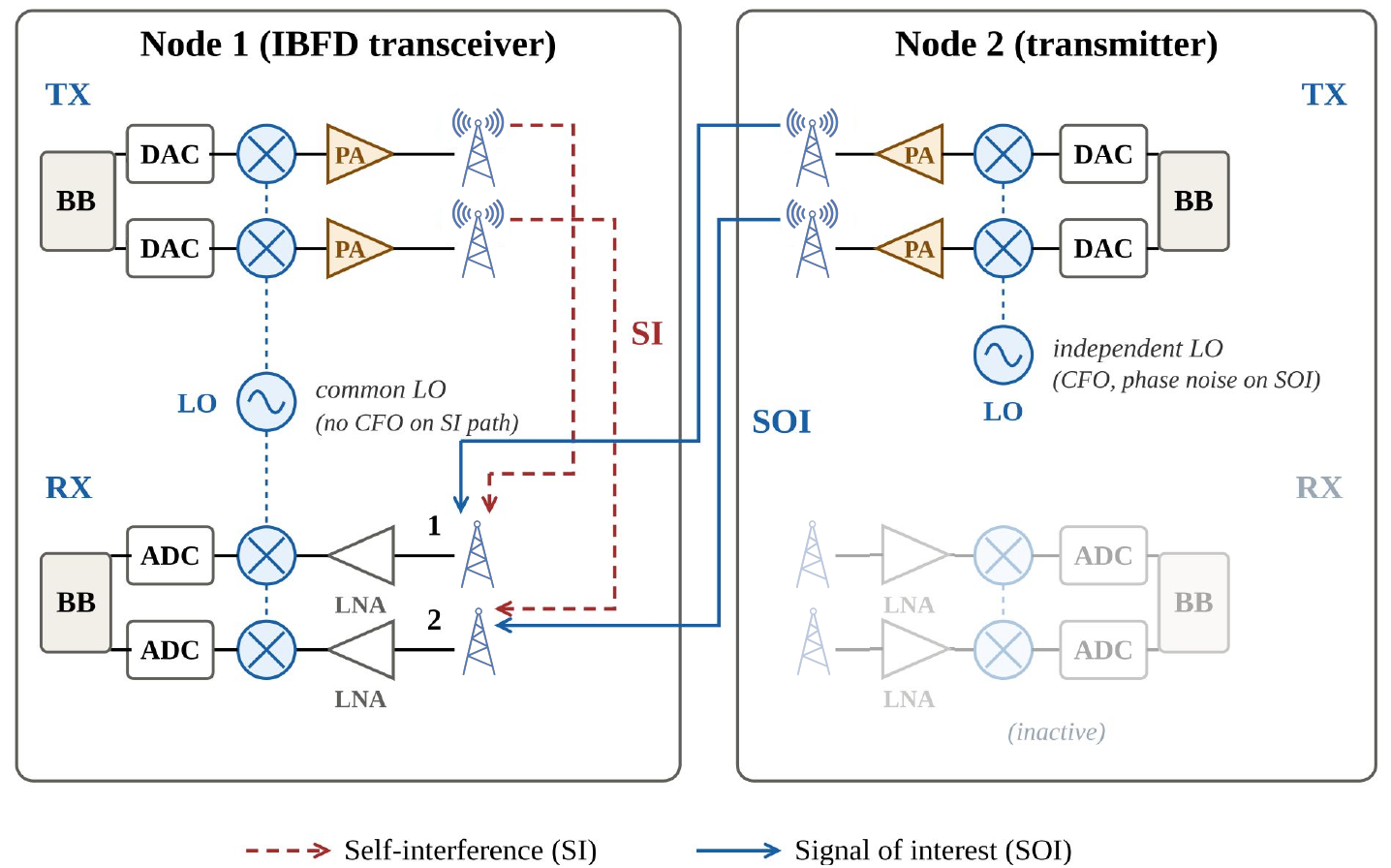}} 
    \caption{2x2 MIMO IBFD system model, showing self-interference (SI) leakage signal and received signal of interest (SOI) at Node 1.}
    \label{fig:System Model}
\end{figure}

This work presents a two-stage ESN-based scheme. 
The architectural contribution is to separate a calibrated SI mapping from a pilot-adapted SOI mapping. 
Since the SI transmitter and receiver reside on the same node (see \figurename~\ref{fig:System Model}), and share a common local oscillator, the SI path is not affected by CFO or phase noise accumulation, unlike the SOI path, which originates from a physically separate transmitter. The SI path’s channel is further assumed static with a fixed antenna geometry between the transmitter and receiver, and the PA nonlinearity is modeled as memoryless. This fixed impairment mapping is characterized offline during an initial calibration phase using a frozen ESN (ESN1). The SOI path, however, exhibits PA nonlinearity with memory and propagation through a time-varying channel, requiring an adaptive ESN (ESN2) to update online via recursive least squares (RLS) using known pilot symbols distributed throughout the transmitted SOI signal.
We evaluate the complete pipeline against the prior Fast-ICA receiver and compare ESN1 followed by ESN2 with the same ESN1 followed by a conventional reservoir-free RLS detector, LSTM, or GRU. Nonlinear SIC methods using the known local transmit reference remain stronger cancellation benchmarks than Fast-ICA \cite{Anttila2013Cancellation,10556632}; their relationship to this evaluation is discussed in Section~\ref{sec:discussion}.
Although applicable to JSAC, this work evaluates communication-side signal recovery; explicit sensing is reserved for future work.

\begin{figure*}[t]
    \centerline{ \includegraphics[width=1.00\linewidth]{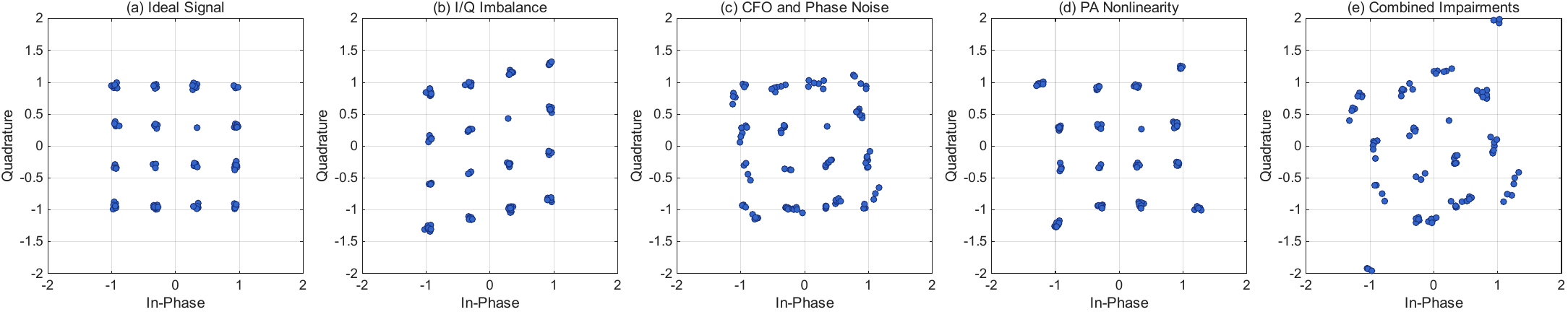}} 
    \caption{RF impairments on a 16-QAM constellation, shown individually and combined.}
    \label{fig:RF Impairments}
\end{figure*}

\label{sec:Related Work}

\section{RF Impairments and Signal Distortion}
\label{sec:Signal Model}

Figure~\ref{fig:RF Impairments} illustrates the individual and combined effects of the modeled RF impairments relative to the ideal received signal
\begin{equation}
\label{equation1}
 \begin{aligned}
    {\bf R}(k) &= {\bf H}(k) {\bf S}(k) + {\bf N}(k),
\end{aligned}
\end{equation}
where \(H(k)\), \(S(k)\), and \(N(k)\) denote the channel, transmitted signal, and AWGN, respectively.

\subsection{I/Q Imbalance}
I/Q imbalance is an impairment that arises from amplitude and phase mismatches between the in-phase and quadrature branches of the transmitter and receiver hardware \cite{Khan2025RFFI}. This mismatch occurs independently at both the transmitter and receiver, since each stage has its own mixers, ADC/DAC converters, and filters, whose imperfections couple the transmitted signal with its own complex conjugate and produce a widely-linear transformation. As shown in \figurename~\ref{fig:RF Impairments}(b), the coupling causes skewed and elliptical distortions in the constellation, shifting it further from the initial symmetrical pattern.

\subsection{CFO and Phase Noise}
CFO and phase noise are impairments caused by imperfections in the crystal oscillator \cite{Wang_CrystalOscillators}. Ideally, the transmitter and receiver oscillators operate at identical frequencies with perfectly stable phase to achieve accurate timing and frequency generation for wireless communication. However, imperfections in manufacturing, crystal aging, and material quality introduce a frequency mismatch, expressed as a net CFO between transmitter and receiver, along with phase instability that accumulates independently at each oscillator as phase noise. As shown in \figurename~\ref{fig:RF Impairments}(c), this causes the constellation to rotate over time due to CFO and jitter randomly due to the combined phase noise from both oscillators.

\subsection{PA Nonlinearity}
Along with the previous impairments, which are linear transformations, there are also nonlinear impairments. PA nonlinearity is a nonlinear transformation caused by the transmitter's PA. An ideal amplifier applies a constant gain, but in practice, the PA's gain compresses as the input amplitude increases and induces a phase shift \cite{Singerl2007Constructing}. These effects are characterized as amplitude/amplitude (AM/AM) and amplitude/phase (AM/PM) distortions where the output amplitude and phase are nonlinear functions of the instantaneous input amplitude. As shown in \figurename~\ref{fig:RF Impairments}(d), this nonlinear compression and phase shift warps the outer symbols of the constellation more severely than the inner ones. This distortion can also exhibit memory, where the output at a specific instant depends on previous samples \cite{Ding2004Memory}.

\subsection{Combined Impaired Signal}

Eq. (\ref{equation2}) summarizes the combined effect of the impairments described above on the received signal. The term $a_1$ represents the linear PA gain, while $\mathbf{H}_1$ and $\mathbf{H}_2$ are the widely-linear coefficient matrices combining the physical channel with the I/Q imbalance-induced conjugate coupling, while $(\cdot)^{*}$ denotes the complex conjugate. The vector $\mathbf{d}(k)$ captures the higher-order nonlinear distortion introduced by PA nonlinearity, corresponding to the $a_3$ and $a_5$ terms. The multiplicative term $e^{j\psi(k)}$ applies the combined phase rotation from CFO and phase noise, where $\Delta f$ is the carrier frequency offset, $T_s$ is the symbol period, and $\phi(k)$ is the accumulated oscillator phase noise. The combined effects of these RF impairments can be visualized by \figurename~\ref{fig:RF Impairments}(e).

\begin{equation}
\label{equation2}
\begin{aligned}
R(k) = {}& e^{j\psi(k)} \Big[ a_1 \mathbf{H}_1 \mathbf{S}(k) + a_1^{*} \mathbf{H}_2 \mathbf{S}^{*}(k) \\
& + \mathbf{H}_1 \mathbf{d}(k) + \mathbf{H}_2 \mathbf{d}^{*}(k) \Big] + \mathbf{N}(k), \\
\end{aligned}
\end{equation}
where $\quad \psi(k) = 2\pi \Delta f k T_s + \phi(k)$.

\section{Echo State Network Pipeline}
\label{sec:ESN}

\begin{figure}[!t]
    \centerline{ \fbox{\includegraphics[width=0.92\linewidth]{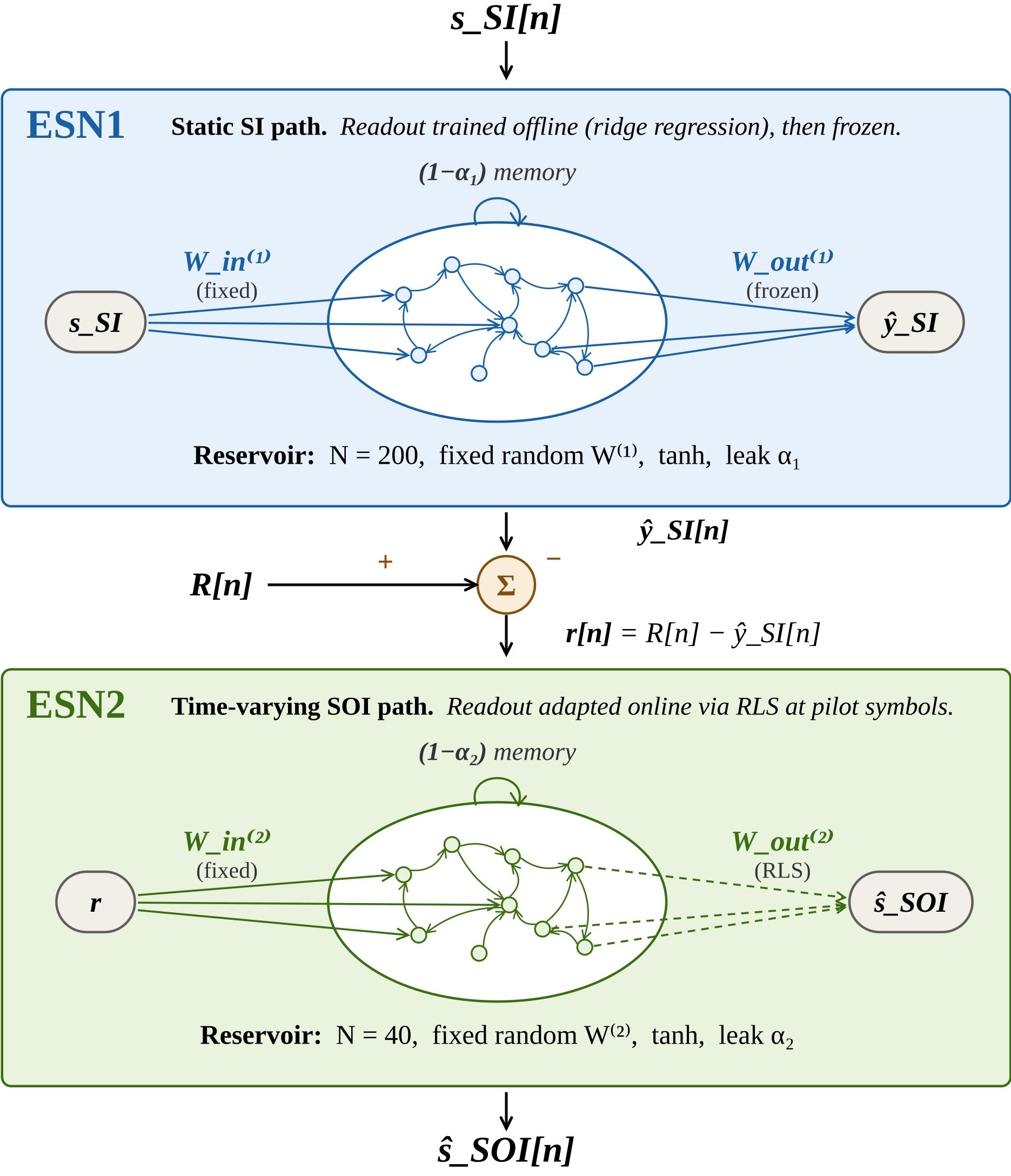}}}
    \caption{Two-stage ESN pipeline for SI cancellation and SOI recovery.}
    \label{fig:ESN Pipeline}
    \vspace{-0.5cm}
\end{figure}

An ESN is a type of RNN composed of three layers: an input layer, the reservoir, and a trained readout layer. The reservoir consists of a fixed number of neurons that are randomly connected to one another, and their weights are initialized once and never updated \cite{9966815}. As an input signal passes through the reservoir at each time index, every neuron's state is updated nonlinearly based on the current input and previous reservoir states. This allows the reservoir to retain short-term memory, capturing time-varying and memory-dependent signal dynamics caused by RF impairments. This high-dimensional reservoir state is mapped to a desired output through the readout layer, which is the only component that is trained. Since the readout layer is a linear combination of the reservoir state, it can be trained through a closed-form solution rather than an iterative, gradient-based solution required by other RNNs. BSS was previously proposed to perform SI suppression and JSAC in IBFD MIMO systems, but its linear, block-based formulation limits performance under realistic RF impairments. To address this, we adopt an ESN architecture to recover the SOI. The two-stage pipeline, as shown in \figurename~\ref{fig:ESN Pipeline}, employs two ESNs: one to estimate the SI signal with impairments for SIC, and another to recover the SOI from the resulting residual $r[n]$. 

\subsection{ESN1}
ESN1 is used to determine the impaired SI signal $y_{{SI}}[n]$ for SIC. Because Node 1 knows its own transmitted SI signal $s_{SI}[n]$, this known signal at index $n$ is used as the input to learn the RF impairments it undergoes after transmission. As shown in Eq. (\ref{eq:x1}), the reservoir state $x_{1}[n]$ is updated at each index using a fixed, randomly initiated input weight matrix $W^{(1)}_{in}$ and reservoir weight matrix $W^{(1)}$, with a leak rate $\alpha_{1}$ that controls how much of the previous reservoir state persists. The estimated impaired SI signal $\hat{y}_{SI}$ is produced by the readout layer, which is trained offline using a ridge regression algorithm during an initial calibration phase. Since the SI path is static and memoryless, as discussed in Section \ref{sec:intro}, ESN1's parameters remain frozen after training, storing a fixed characterization of the RF impairments that can be applied for SIC during live transmission from both nodes.

\begin{equation}
\begin{aligned}
x_{1}[n] ={}& (1-\alpha_{1})x_{1}[n-1] \\
&+ \alpha_{1}\tanh\left(
W_{in}^{(1)}s_{SI}[n]
+ W^{(1)}x_{1}[n-1]
\right).
\end{aligned}
\label{eq:x1}
\end{equation}

\subsection{ESN2}
ESN2 recovers $\hat{s}_{SOI}[n]$ from the residual $r[n] = R[n]-\hat{y}_{SI}[n]$. Unlike ESN1, the transmitted SOI signal is unknown to Node 1, since it originates from a physically separate transmitter, Node 2. To recover it, a small fraction of known pilot symbols are distributed throughout the transmitted SOI signal, providing a ground-truth target at sparse intervals. As shown in Eq. (\ref{eq:x2}), the reservoir state $x_2[n]$ is updated at each index using a fixed, randomly initialized input weight matrix $W_{in}^{(2)}$ and reservoir weight matrix $W^{(2)}$, with a leak rate $\alpha_2$ that controls how much of the previous reservoir state persists. Because the SOI path exhibits PA nonlinearity with memory and propagates through a time-varying channel, as discussed in Section \ref{sec:intro}, a fixed offline-trained readout, as used in ESN1, is insufficient to track its changing statistics. Instead, ESN2's readout is updated online via recursive least squares (RLS) at each pilot index, allowing its parameters to continuously adapt to the evolving SOI channel using only the sparse pilot symbols available during live transmission.

\begin{equation}
\begin{aligned}
x_{2}[n] ={}& (1-\alpha_{2})x_{2}[n-1] \\
&+ \alpha_{2}\tanh\left(
W_{in}^{(2)}r[n]
+ W^{(2)}x_{2}[n-1]
\right).
\end{aligned}
\label{eq:x2}
\end{equation}

\section{Simulation}
\label{sec:simulation}

We simulate an IBFD MIMO system under the physical setup and RF impairments described in Section \ref{sec:Signal Model}. We assess SOI recovery performance against the Fast-ICA method from our prior work, compare the proposed ESN2 against LSTM and GRU baselines under equivalent online adaptation, and evaluate the pipeline's acquisition speed across a range of block sizes.

\subsection{ESN1 and ESN2 Setup}
\label{sec:esnsetup}
We evaluated the proposed two-stage ESN pipeline through simulations. Both the SI and SOI signals are generated with equal power levels using 16-QAM symbols, with 2 transmit and receive antennas at Node 1, and 2 transmit antennas at Node 2, with a signal-to-noise ratio (SNR) set to 20 dB. At the input to the digital canceller, the SI-to-SOI power ratio is $4.58$~dB, measured after the respective channels and RF impairments and before AWGN is added. The reservoir size, spectral radius, input scaling, leak rate, and regularization parameter are hyperparameter-tuned using a grid search algorithm for both ESN1 and ESN2.

As mentioned in Section \ref{sec:ESN}, ESN1 is trained offline using 500 blocks of 500 symbols each. The data is split 70/30 into training and test sets with a reservoir size of 200 neurons, spectral radius of 0.95, input scaling of 0.5, leak rate $\alpha_1$ of 0.5, ridge regression regularization parameter of $10^{-6}$, and a washout period of 50 samples. 
Once ESN1 is trained, the frozen parameters are used to estimate the impaired SI signal $\hat{y}_{SI}[n]$ which is subtracted from the received signal $R[n]$ to result in the residual $r[n]$ for SIC. This residual serves as the input to ESN2. The SOI path is modeled as a controlled linearly time-varying three-tap MIMO channel, with a symbol-wise variation rate of \(\delta=10^{-4}\) in the diagonal coefficients and fixed delayed matrices at one- and two-symbol delays.
The SOI signal is affected by CFO set to 100 Hz and phase noise, and a memory polynomial PA nonlinearity with 3 delay taps and an exponentially decaying coefficient factor of 0.8. The complete set of RF impairment and channel parameters for both the SI and SOI paths is listed in Table \ref{tab:rf_params}. To track these adaptive dynamics, ESN2 uses a smaller reservoir size of 40 neurons, a spectral radius of 0.9, an input scaling of 0.1, and a leak rate $\alpha_2$ of 0.9. The readout is updated online via RLS with a forgetting factor of 0.998 and initial regularization of 1. The pilot symbols make up 15\% of the total block size distributed throughout the transmitted SOI signal for online adaptation.

\begin{table}[!ht]
\centering
\caption{RF Impairment and Channel Parameters}
\label{tab:rf_params}
\resizebox{0.49\textwidth}{!}{%
\begin{tabular}{lcc}
\toprule
\textbf{Parameter} & \textbf{SI Path} & \textbf{SOI Path} \\
\midrule
Phase noise std. dev. & $0.05^\circ$ & $0.05^\circ$ \\
TX amplitude imbalance $\epsilon_{tx}$ & $0.02$ & $0.02$ \\
TX phase imbalance $\phi_{tx}$ & $2^\circ$ & $2^\circ$ \\
RX amplitude imbalance $\epsilon_{rx}$ & $0.01$ & $0.01$ \\
RX phase imbalance $\phi_{rx}$ & $1^\circ$ & $1^\circ$ \\
CFO & $0$ Hz & $100$ Hz \\
Phase rotation (CFO/PN) & Disabled & Enabled \\
\midrule
PA taps $Q$ & 1 (memoryless) & 3 \\
$a_1$ (per tap) & $1.0$ & $[1.0,\ 0.20,\ 0.08]$ \\
$a_3$ (per tap) & $-0.18 - 0.03j$ & $[-0.140,\ -0.112,\ -0.090] - j[0.015,\ 0.012,\ 0.010]$ \\
$a_5$ (per tap) & $0.15 + 0.06j$ & $[0.080,\ 0.064,\ 0.051] + j[0.030,\ 0.024,\ 0.019]$ \\
\bottomrule
\end{tabular}
}
\end{table}

\subsection{BSS and ESN Pipeline Setup}
To compare against our prior work, the BSS algorithm, Fast-ICA, is evaluated under the same RF impairments described in Section \ref{sec:esnsetup} and the values used in Table \ref{tab:rf_params}. To achieve the highest performance for the ESN pipeline in recovering the SOI, the target BER is set to $10^{-2}$; otherwise, the ESN pipeline would terminate prematurely at a looser threshold. Both models are evaluated by averaging 200 Monte Carlo trials with varying block sizes from 50 to 500, incrementing by 50.

\subsection{Comparative Models and ESN2}
Since ESN1 remains frozen after training, our comparison against alternative RNNs, as well as an additional RLS-only baseline, focuses on the adaptive stage represented by ESN2. The RLS baseline applies the same RLS-updated readout directly to $r[n]$ without a reservoir, evaluating the reservoir’s joint nonlinear-state and fading-memory contribution.
LSTM and GRU baselines \cite{LSTM, GRU} are sized to match ESN2's total trainable parameter count, using hidden sizes of 19 and 22, respectively, to ensure a fair comparison. The LSTM and GRU baselines are adapted online through gradient-based updates at each pilot symbol with a learning rate of $10^{-2}$, and all three models use the same input and output, mapping the residual $r[n]$ to the estimated SOI $\hat{s}_{SOI}[n]$.

Because the LSTM and GRU baselines' gradient-based online updates converge slower than ESN2's closed-form RLS readout, a target BER of $10^{-2}$ prevented some trials from converging within the maximum number of processed blocks of 500. For this comparison, the target BER is set to $10^{-1}$, ensuring all models reach a consistent criterion within the trial limit and allowing a fair comparison of efficiency across methods. Each model is evaluated by averaging the results of 50 individual Monte Carlo trials with varying block sizes from 50 to 500, incrementing by 50. For each trial, the receiver is considered acquired once the target BER is sustained for three consecutive fresh blocks of data, up to a maximum of 500.
We compare the resulting SOI recovery SRER and the average number of blocks required to converge across all tested block sizes to measure performance and efficiency.

\section{Results and Discussion}
\label{sec:results}


\subsection{ESN vs BSS (Fast-ICA)}

\begin{figure}[!t]
    \centerline{{\includegraphics[width=.92\linewidth]{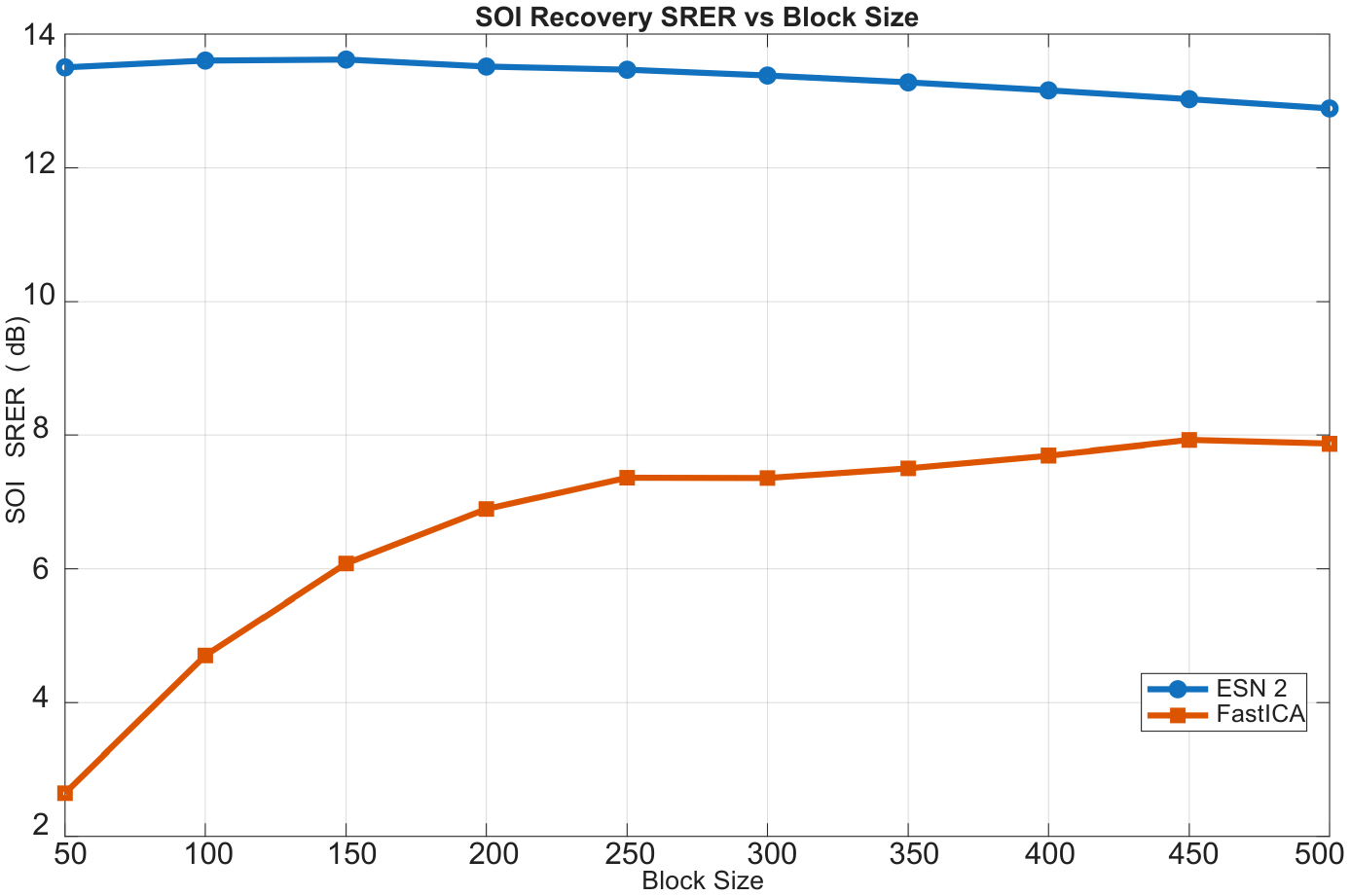}}}
    \caption{ESN vs BSS (Fast-ICA) for SOI recovery SRER.}
    \label{fig:ESNFICAI}
    \vspace{-0.5cm}
\end{figure}

As shown in \figurename~\ref{fig:ESNFICAI}, the ESN pipeline outperforms Fast-ICA in SOI recovery SRER in decibels across all tested block sizes, with the largest gap of approximately 11 dB occurring at a block size of 50, where Fast-ICA's linear, block-based estimation is most limited. As block size increases, ESN performance degrades by less than 1 dB, since more samples elapse per block, allowing the CFO and phase noise to accumulate further before ESN2's next pilot update. Fast-ICA, on the other hand, considers the whole block of data at once for each estimate, so it is not affected by this accumulation across blocks. 

\subsection{ESN2 vs RLS/LSTM/GRU}

\begin{figure}[!t]
\centering
\begin{subfigure}[b]{\linewidth}
    \centering
    \includegraphics[width=0.92\linewidth]{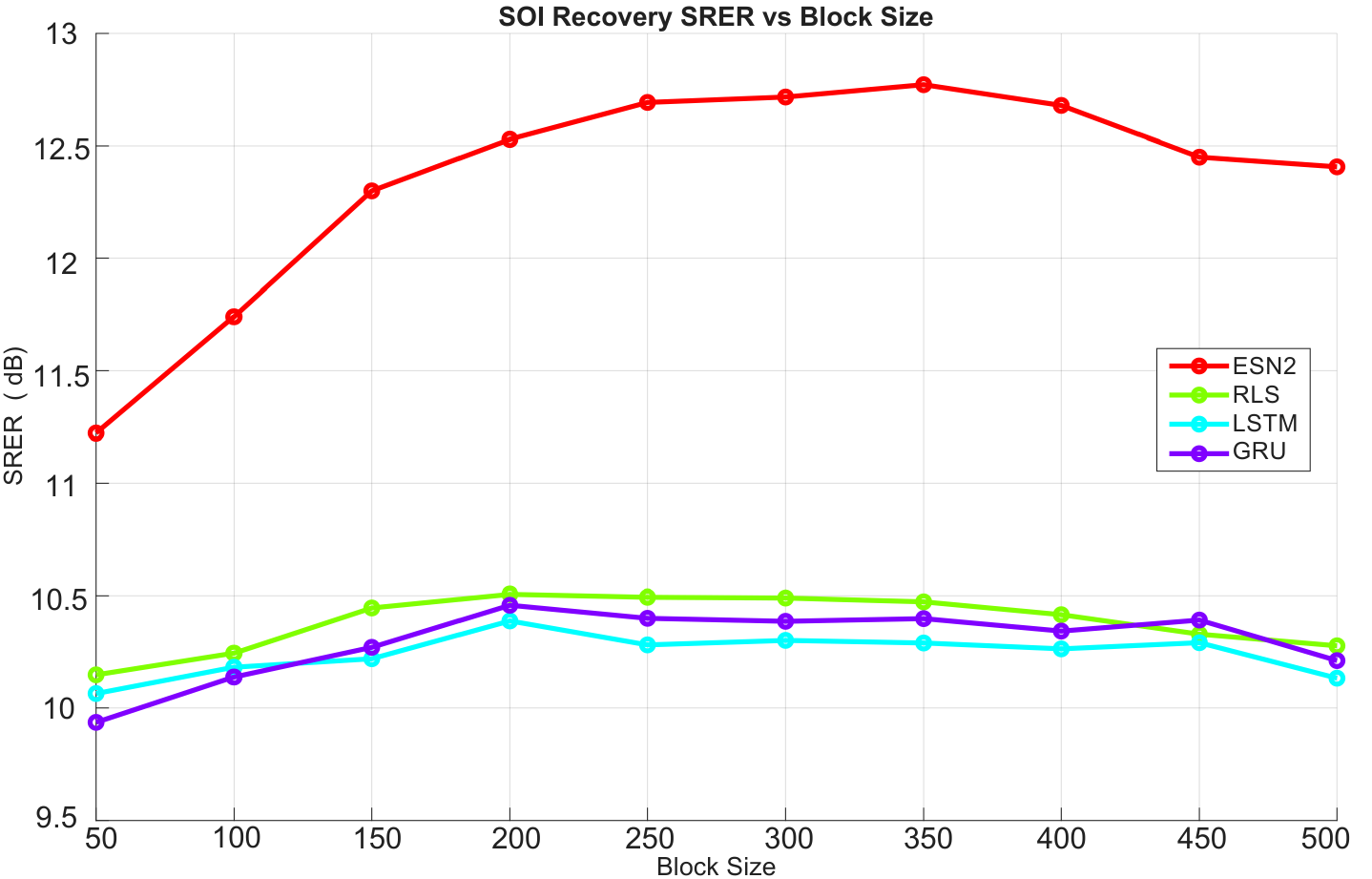}
    \caption{SOI recovery SRER comparison.}
    \label{fig:ESNSOI}
\end{subfigure}
\begin{subfigure}[b]{\linewidth}
    \centering
    \includegraphics[width=0.92\linewidth]{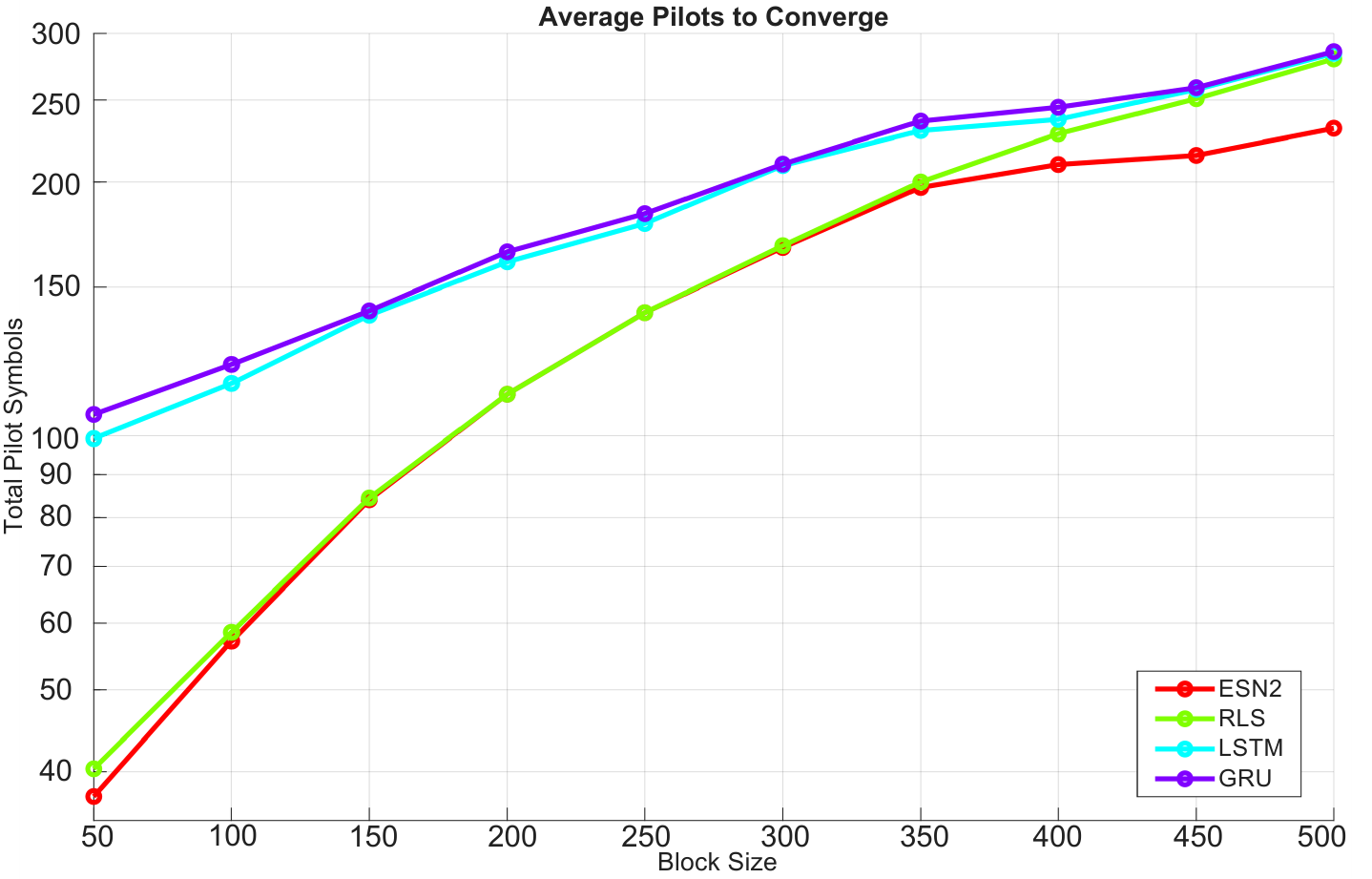}
    \caption{Pilots required to converge to target BER.}
    \label{fig:ESNvsBaseline}
\end{subfigure}
\caption{ESN2 vs RLS/LSTM/GRU: recovery accuracy and pilot efficiency.}
\label{fig:ESNComparison}
\vspace{-0.5cm}
\end{figure}

\figurename~\ref{fig:ESNSOI} compares the SOI recovery SRER of ESN2 against RLS, LSTM, and GRU. ESN2 achieves the highest SRER across all tested block sizes, achieving approximately 13 dB. As mentioned in Section \ref{sec:simulation}, the results between \figurename~\ref{fig:ESNFICAI} and \figurename~\ref{fig:ESNSOI} differ because the target BER is relaxed for this comparison to allow all the models to converge for accurate comparison. RLS, LSTM, and GRU results all cluster closely between 9.5 and 10.5 dB. Since RLS shares ESN2's closed-form readout but performs comparably to the gradient-based LSTM and GRU rather than approaching ESN2, this result attributes the improvement to the reservoir, including its nonlinear state expansion and fading-memory dynamics.


\figurename~\ref{fig:ESNvsBaseline} compares the average number of pilot symbols required for ESN2, RLS, LSTM, and GRU to converge. At small block sizes, ESN2 and RLS require a low number of pilots, well below LSTM and GRU. As block size increases, however, RLS's pilot requirement rises towards LSTM and GRU, while ESN2 remains the most pilot-efficient method throughout, requiring approximately 230 pilot symbols at a block size of 500, compared to approximately 280$-$290 for RLS, LSTM, and GRU. This indicates that ESN2's reservoir not only contributes to higher recovery accuracy, but also to maintaining pilot efficiency at larger block sizes, an advantage the RLS  baseline does not retain.

\subsection{Discussion and Computational Complexity}
\label{sec:discussion}
Fast-ICA is a continuity baseline for our earlier BSS receiver~\cite{Li2025MILCOM, Li2025Electronics}, not a state-of-the-art nonlinear SI canceller. It estimates a linear mixture, whereas reference-based SIC exploits the known local transmit sequence. Anttila et al.\ \cite{Anttila2013Cancellation} model nonlinear SI using a parallel Hammerstein structure, accounting for PA distortion and memory. Liu et al.\ \cite{10556632} instead learn the SI mapping with RC and benchmark it against polynomial and neural-network SIC. That work evaluates cancellation depth and computational demand, while our figures evaluate post-cancellation SOI SRER and pilot acquisition. These metrics and setups do not support a numerical ranking against \cite{10556632}.

Among the adaptive baselines, the RLS model shares ESN2's closed-form readout but does not include the reservoir. RLS's low pilot cost at small block sizes shows that the closed-form update alone drives early acquisition speed, while its lower SRER and rising pilot cost at larger block sizes show that the reservoir improves both recovery accuracy and efficiency. LSTM and GRU achieve SRER comparable to RLS, but without RLS's early pilot efficiency, since their iterative, gradient-based updates require more data to converge than a closed-form update, reflecting both the absence of a reservoir and the lack of a closed-form readout. Together, these results support the motivation of this work by combining a fixed reservoir with a closed-form adaptive readout, which enables reliable, real-time SOI recovery under simulated RF impairments and time-varying channel dynamics.

We measure computational complexity by the number of multiplications per input symbol. ESN1 requires $42020$ multiplications because of its $200$ neuron reservoir, but it is trained offline and remains frozen during operation. In contrast, the online-adaptive ESN2, RLS, LSTM, and GRU are subject to real-time constraints. ESN2 requires $2020$ multiplications per symbol, compared with $2064$ for LSTM and $2028$ for GRU.
Since ESN2's readout is updated via a single closed-form RLS step per pilot, it requires no gradient computation, contributing to fewer pilots required to reach the target BER than the gradient-based LSTM and GRU baselines, consistent with the pilot-efficiency results shown in \figurename~\ref{fig:ESNvsBaseline}.

\section{Conclusion and Future Work}
\label{sec:conclusion}
We investigated an IBFD MIMO receiver and 
 proposed a two-stage ESN pipeline, using a frozen ESN to characterize the static SI path offline and an adaptive ESN, updated online via RLS, to track the time-varying SOI path. The existing results support improved SOI recovery over Fast-ICA and, with ESN1 fixed, improved recovery and pilot efficiency over the tested RLS, LSTM, and GRU detectors. The detector ablation measures the joint contribution of nonlinear state expansion and fading memory. 
 Future work will extend the proposed ESN framework toward JSAC and physical-layer security by acquisition techniques developed for downlink JSAC~\cite{Song2024}, reservoir states as hardware features for RF fingerprinting, and learned RF impairments for spoofing detection.



\bibliographystyle{IEEEtran}
\bibliography{ref}

@article{heath2016overview,
  title={An overview of signal processing techniques for millimeter wave MIMO systems},
  author={Heath, Robert W and Gonzalez-Prelcic, Nuria and Rangan, Sundeep and Roh, Wonil and Sayeed, Akbar M},
  journal={IEEE journal of selected topics in signal processing},
  volume={10},
  number={3},
  pages={436--453},
  year={2016},
  publisher={IEEE}
}

@book{kolodziej2021band,
  title={In-Band Full-Duplex Wireless Systems Handbook},
  author={Kolodziej, Kenneth E},
  year={2021},
  publisher={Artech House}
}

@book{alves2020full,
  title={Full-duplex communications for future wireless networks},
  author={Alves, Hirley and Riihonen, Taneli and Suraweera, Himal A},
  year={2020},
  publisher={Springer}
}

@INPROCEEDINGS{Li2025MILCOM,
  author={Li, Siyao and Prisby, Conrad and Yang, Thomas},
  booktitle={MILCOM 2025 - 2025 IEEE Military Communications Conference (MILCOM)},
  title={Blind Source Separation-Enabled Joint Communication and Sensing in IBFD MIMO Systems},
  year={2025},
  month={Oct.},
  pages={1223-1228},
  doi={10.1109/MILCOM64451.2025.11309844}
}

@article{Li2025Electronics,
  author={Li, Siyao and Prisby, Conrad and Yang, Thomas},
  title={Blind Source Separation for Joint Communication and Sensing in Time-Varying IBFD MIMO Systems},
  journal={Electronics},
  year={2025},
  month={Jan.},
  volume={14},
  number={16},
  pages={3200},
  doi={10.3390/electronics14163200}
}

@misc{Khan2025RFFI,
  title={A Comprehensive Survey on Feature Extraction Techniques Using I/Q Imbalance in RFFI},
  author={Khan, M. A. and Siddiqui, M. U.},
  year={2025},
  month={Feb.},
  eprint={2502.02782},
  archivePrefix={arXiv},
  doi={10.48550/arXiv.2502.02782}
}

@article{Wang_CrystalOscillators,
  author={Wang, J. and Liu, X.},
  title={A Review of Crystal Oscillators Imperfection: Linking Frequency Deviations to Carrier Frequency Offset and Phase Noise},
  journal={Journal of Networking and Network Applications},
  year={2024},
  month={Dec.},
  volume={4},
  number={4},
  pages={165-171},
  doi={https://doi.org/10.33969/J-NaNA.2024.040403}
}

@INPROCEEDINGS{Anttila2013Cancellation,
  author={Anttila, L. and Korpi, D. and Syrjala, V. and Valkama, M.},
  booktitle={2013 Asilomar Conference on Signals, Systems and Computers},
  title={Cancellation of power amplifier induced nonlinear self-interference in full-duplex transceivers},
  year={2013},
  month={Nov.},
  address={Pacific Grove, CA},
  publisher={IEEE},
  pages={1193-1198},
  doi={10.1109/ACSSC.2013.6810482}
}

@article{Ding2004Memory,
  author={Ding, L. and Zhou, G.T. and Morgan, D.R. and Ma, Z. and Kenney, J.S. and Kim, J.},
  journal={IEEE Transactions on Communications},
  title={A robust digital baseband predistorter constructed using memory polynomials},
  year={2004},
  month={Jan.},
  volume={52},
  number={1},
  pages={159-165},
  doi={10.1109/TCOMM.2003.822188}
}

@INPROCEEDINGS{Singerl2007Constructing,
  author={Singerl, Peter and Kubin, Gernot},
  booktitle={2007 50th Midwest Symposium on Circuits and Systems}, 
  title={Constructing memory-polynomial models from frequency-dependent AM/AM and AM/PM measurements}, 
  year={2007},
  volume={},
  number={},
  pages={321-324},
  doi={10.1109/MWSCAS.2007.4488598}
  }

@article{li2023real,
  title={Real-time machine learning for multi-user massive MIMO: Symbol detection using multi-mode StructNet},
  author={Li, Lianjun and Xu, Jiarui and Zheng, Lizhong and Liu, Lingjia},
  journal={IEEE Transactions on Wireless Communications},
  volume={22},
  number={12},
  pages={9172--9186},
  year={2023},
  publisher={IEEE}
}

@INPROCEEDINGS{Li2022,
  author={Li, Siyao and Caire, Giuseppe},
  booktitle={2022 56th Asilomar Conference on Signals, Systems, and Computers}, 
  title={On the Capacity of “Beam-Pointing” Channels with Block Memory and Feedback: The Binary Case}, 
  year={2022},
  volume={},
  number={},
  pages={1262-1268},
  doi={10.1109/IEEECONF56349.2022.10051895}}

@INPROCEEDINGS{Li2023,
  author={Li, Siyao and Caire, Giuseppe},
  booktitle={2023 IEEE International Symposium on Information Theory (ISIT)}, 
  title={On the Capacity and State Estimation Error of Binary "Beam-Pointing" Channels with Block Memory and Feedback}, 
  year={2023},
  volume={},
  number={},
  pages={2571-2576},
  doi={10.1109/ISIT54713.2023.10206568}}

@INPROCEEDINGS{Li2024,
  author={Li, Siyao and Pedraza, Fernando and Caire, Giuseppe},
  booktitle={2024 IEEE International Symposium on Information Theory (ISIT)}, 
  title={On the Capacity of Gaussian “Beam-Pointing” Channels with Block Memory and Feedback}, 
  year={2024},
  volume={},
  number={},
  pages={2371-2376},
  doi={10.1109/ISIT57864.2024.10619226}}

@INPROCEEDINGS{Song2024,
  author={Song, Yi and Pedraza, Fernando and Li, Shuangyang and Li, Siyao and Yu, Han and Caire, Giuseppe},
  booktitle={ICC 2024 - IEEE International Conference on Communications}, 
  title={Compressed Sensing Inspired User Acquisition for Downlink Integrated Sensing and Communication Transmissions}, 
  year={2024},
  volume={},
  number={},
  pages={5293-5298},
  doi={10.1109/ICC51166.2024.10622469}}

@ARTICLE{9966815,
  author={Sun, Chenxi and Song, Moxian and Cai, Derun and Zhang, Baofeng and Hong, Shenda and Li, Hongyan},
  journal={IEEE Transactions on Artificial Intelligence}, 
  title={A Systematic Review of Echo State Networks From Design to Application}, 
  year={2024},
  volume={5},
  number={1},
  pages={23-37},
  doi={10.1109/TAI.2022.3225780}}

@ARTICLE{8169663,
  author={Mosleh, Somayeh Susanna and Liu, Lingjia and Sahin, Cenk and Zheng, Yahong Rosa and Yi, Yang},
  journal={IEEE Transactions on Neural Networks and Learning Systems}, 
  title={Brain-Inspired Wireless Communications: Where Reservoir Computing Meets MIMO-OFDM}, 
  year={2018},
  volume={29},
  number={10},
  pages={4694-4708},
  doi={10.1109/TNNLS.2017.2766162}}

@ARTICLE{LSTM,
  author={Hochreiter, Sepp and Schmidhuber, Jürgen},
  journal={Neural Computation}, 
  title={Long Short-Term Memory}, 
  year={1997},
  volume={9},
  number={8},
  pages={1735-1780},
  doi={10.1162/neco.1997.9.8.1735}}

@inproceedings{GRU,
  author    = {Junyoung Chung and
               {\c{C}}aglar G{\"u}l{\c{c}}ehre and
               KyungHyun Cho and
               Yoshua Bengio},
  title     = {Empirical Evaluation of Gated Recurrent Neural Networks on Sequence Modeling},
  booktitle = {NIPS 2014 Workshop on Deep Learning and Representation Learning},
  address   = {Montreal, Canada},
  year      = {2014}
}

@ARTICLE{10556632,
  author={Liu, Zhikai and Luo, Haifeng and Ratnarajah, Tharmalingam},
  journal={IEEE Transactions on Machine Learning in Communications and Networking}, 
  title={Reservoir Computing-Based Digital Self-Interference Cancellation for In-Band Full-Duplex Radios}, 
  year={2024},
  volume={2},
  number={},
  pages={855-868},
  doi={10.1109/TMLCN.2024.3414296}}
\end{document}